\pdfoutput=1

\documentclass[11pt]{article}

\usepackage{amsmath}
\usepackage{tikz}
\usepackage{subcaption}
\expandafter\def\csname ver@subfig.sty\endcsname{}
\usepackage{xcolor}
\usepackage{bm}
\usepackage[hyphens]{url}
\usepackage{acl}
\usepackage{todonotes}
\usepackage{times}
\usepackage{latexsym}
\usepackage{subfig}
\usepackage{listings}
\usepackage{tcolorbox}
\usepackage{siunitx}
\usepackage{enumitem}
\usepackage[T1]{fontenc}

\usepackage[utf8]{inputenc}

\usepackage{microtype}

\usepackage{inconsolata}

\usepackage{graphicx}

\title{Defining bias in AI-systems: Biased models are fair models}

\author{Chiara Lindloff \and Ingo Siegert \\
        Mobile Dialog Systems Group \\ Otto von Guericke University Magdeburg, Germany}

\begin{document}
\maketitle
\begin{abstract}
The debate around bias in AI systems is central to discussions on algorithmic fairness. However, the term “bias” often lacks a clear definition, despite frequently being contrasted with “fairness” — implying that an unbiased model is inherently fair. In this paper, we challenge this assumption and argue that a precise conceptualization of bias is necessary to effectively address fairness concerns. Rather than viewing bias as inherently negative or unfair, we highlight the importance of distinguishing between bias and discrimination. We further explore how this shift in focus can foster a more constructive discourse within academic debates on fairness in AI systems.
\end{abstract}

\section{Introduction}
Bias mitigation is a key topic in current discussions surrounding AI systems~\cite{gray2024bias}. 
But what exactly is bias? From a technical perspective, bias describes input modulation in neural networks. However, beyond this technical definition lies a socially relevant dimension that necessitates mitigation strategies. Bias is often understood as a systematic error in judgment, rooted in prejudice, that perpetuates existing power structures such as racism or sexism—making outcomes unfair. Consequently, mitigation strategies, aimed at promoting fairness, typically focus on eliminating bias by equalizing treatment across groups~\cite{bell2015bias,bender2021stochastic,ibm2024aibias,kartal2022bias,ntoutsi2020bias}.

Recent research highlights the complexity and importance of bias mitigation. For example, Berkeley Haas School of Business provides practical strategies for identifying and mitigating bias to promote responsible and equitable use of AI~\cite{berkeley2024playbook}. The National Institute of Standards and Technology (NIST) emphasizes the need for a socio-technical approach to address bias and highlights the role of context in mitigation strategies~\cite{nist2023bias}. Additionally, a large-scale empirical study evaluates 17 different bias mitigation methods, providing insights into their effectiveness and impact on fairness~\cite{10.1145/3583561}. These examples illustrate the range of approaches in current bias mitigation research—from practical frameworks to empirical evaluations.

Yet, what does it actually mean to say a model is biased? Despite its frequent juxtaposition with “fairness”, the term remains diffuse and ambiguously defined. This paper seeks to clarify what bias truly is, why it requires mitigation, and how its conceptualization shapes the discourse on fairness in AI systems.

\section{What 'Bias' Means in Neural Networks: A Look at Its History and Technical Role}

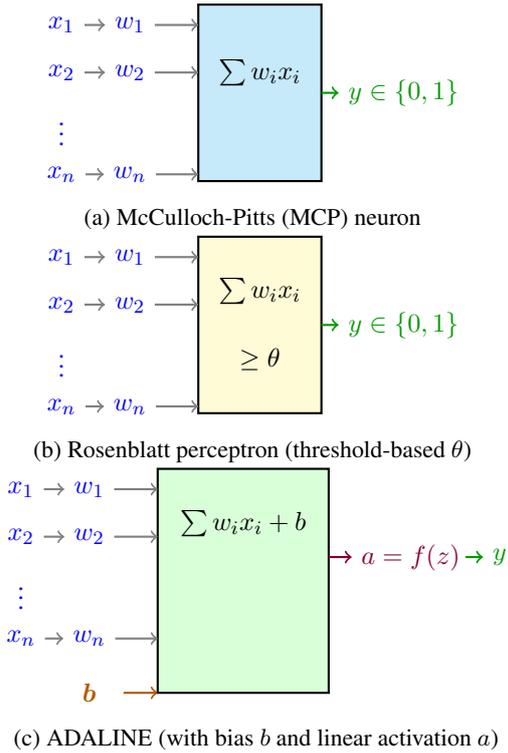
\begin{figure}[h]
    \centering
    \begin{subfigure}[b]{0.45\textwidth}
    \centering
    \begin{tikzpicture}[thick,scale=0.9, every node/.style={scale=0.9}]
        \node[blue] at (-2,1) (x1) {$x_1$};
        \node[blue] at (-2,0.3) (x2) {$x_2$};
        \node[blue] at (-2,-0.5) (dots) {$\vdots$};
        \node[blue] at (-2,-1.2) (xn) {$x_n$};

        \node[blue] at (-1,1) (w1) {$w_1$};
        \node[blue] at (-1,0.3) (w2) {$w_2$};
        \node[blue] at (-1,-1.2) (wn) {$w_n$};

        \fill[cyan!20] (0,-1.3) rectangle (1.8,1.3);
        \draw[thick] (0,-1.3) rectangle (1.8,1.3);
        \node at (0.9,0.3) {$\sum w_i x_i$};

        \draw[->,gray,thick] (x1) -- (w1);
        \draw[->,gray,thick] (x2) -- (w2);
        \draw[->,gray,thick] (xn) -- (wn);

        \draw[->,gray,thick] (w1) -- (0,1);
        \draw[->,gray,thick] (w2) -- (0,0.3);
        \draw[->,gray,thick] (wn) -- (0,-1.2);

        \node[green!60!black] at (3,0) (y) {$y \in \{0,1\}$};
        \draw[->,green!60!black,thick] (1.8,0) -- (y);
    \end{tikzpicture}
    \caption{McCulloch-Pitts (MCP) neuron}
    \label{fig:MCP}
    \end{subfigure}
    \hfill
    
    \begin{subfigure}[b]{0.45\textwidth}
    \centering
    \begin{tikzpicture}[thick,scale=0.9, every node/.style={scale=0.9}]
        \node[blue] at (-2,1) (x1) {$x_1$};
        \node[blue] at (-2,0.3) (x2) {$x_2$};
        \node[blue] at (-2,-0.5) (dots) {$\vdots$};
        \node[blue] at (-2,-1.2) (xn) {$x_n$};

        \node[blue] at (-1,1) (w1) {$w_1$};
        \node[blue] at (-1,0.3) (w2) {$w_2$};
        \node[blue] at (-1,-1.2) (wn) {$w_n$};

        \fill[yellow!20] (0,-1.3) rectangle (1.8,1.3);
        \draw[thick] (0,-1.3) rectangle (1.8,1.3);
        \node at (0.9,0.5) {$\sum w_i x_i$};
        \node at (0.9,-0.5) {$\geq \theta$};

        \draw[->,gray,thick] (x1) -- (w1);
        \draw[->,gray,thick] (x2) -- (w2);
        \draw[->,gray,thick] (xn) -- (wn);

        \draw[->,gray,thick] (w1) -- (0,1);
        \draw[->,gray,thick] (w2) -- (0,0.3);
        \draw[->,gray,thick] (wn) -- (0,-1.2);

        \node[green!60!black] at (3,0) (y) {$y \in \{0,1\}$};
        \draw[->,green!60!black,thick] (1.8,0) -- (y);
    \end{tikzpicture}
    \caption{Rosenblatt perceptron (threshold-based $\theta$)}
    \label{fig:Rosenblatt}
    \end{subfigure}
    \hfill
    
    \begin{subfigure}[b]{0.45\textwidth}
    \centering
    \begin{tikzpicture}[thick,scale=0.9, every node/.style={scale=0.9}]
        \node[blue] at (-2,1) (x1) {$x_1$};
        \node[blue] at (-2,0.3) (x2) {$x_2$};
        \node[blue] at (-2,-0.5) (dots) {$\vdots$};
        \node[blue] at (-2,-1.2) (xn) {$x_n$};

        \node[blue] at (-1,1) (w1) {$w_1$};
        \node[blue] at (-1,0.3) (w2) {$w_2$};
        \node[blue] at (-1,-1.2) (wn) {$w_n$};

        \node[orange!70!black] at (-1,-2) (b) {$\bm{b}$};
        \draw[->,orange!70!black,thick] (-0.5,-2) -- (0,-2);

        \fill[green!15] (0,-2) rectangle (2.5,1.3);
        \draw[thick] (0,-2) rectangle (2.5,1.3);
        \node at (1.25,0.5) {$\sum w_i x_i + b$};

        \draw[->,gray,thick] (x1) -- (w1);
        \draw[->,gray,thick] (x2) -- (w2);
        \draw[->,gray,thick] (xn) -- (wn);

        \draw[->,gray,thick] (w1) -- (0,1);
        \draw[->,gray,thick] (w2) -- (0,0.3);
        \draw[->,gray,thick] (wn) -- (0,-1.2);

        \node[purple!70!black] at (3.7,0) (act) {$a = f(z)$};
        \draw[->,purple!70!black,thick] (2.5,0) -- (act);

        \node[green!60!black] at (5,0) (y) {$y$};
        \draw[->,green!60!black,thick] (4.5,0) -- (y);
    \end{tikzpicture}
    \caption{ADALINE (with bias $b$ and linear activation $a$)}
    \label{fig:Adaline}
    \end{subfigure}
    \caption{Comparison of discussed early neuron models.}
\end{figure}

The elementary building blocks of neural networks, artificial mechanical neurons, trace their origin to the founding fathers of the technology that now builds our large language models (LLMs) — they may also be referred to as MCP neurons, or McCulloch-Pitts-Neurons~\cite{mcculloch1943logical}, which are simple binary neuron models that process inputs using a threshold value, account for inhibitory signals, and produce an output of either 0 or 1~\cite{vijaychandra2019mcp}. Figure~\ref{fig:MCP} illustrates the basic principles of an MCP.

This concept was further developed by Frank Rosenblatt, whose perceptron model, see Figure~\ref{fig:Rosenblatt}, paved the way for modern machine learning algorithms by introducing adjustable weights and a learning rule for updating these weights based on errors~\cite{rosenblatt1962perceptrons}. Although the perceptron relied on a threshold for classification, the explicit concept of a bias term to shift the decision boundary was not yet introduced~\cite{vijaychandra2019mcp} Its adaptive capabilities laid the groundwork for pretty much all modern algorithmic solutions, as its main goal is to make accurate classification. The basic functionality allows for linearly separable classification according to whether or not the weighted inputs meet a certain threshold $\theta$. 

In 1960, Bernard Widrow built on Rosenblatt’s perceptron and developed ADALINE~\cite{widrow1988layered}, which is where the term ‘bias’ was first explicitly used to denote a constant input node that adjusts the model’s decision boundary independently of the input values, see Figure\ref{fig:Adaline}. ADALINE (Adaptive Linear Neuron) is a single-layer neural network that uses a linear activation function and applies the Least Mean Squares (LMS) learning rule to minimize the error between predicted and actual outputs during training. The model was implemented in a physical device called Memistor, which simulated the learning process by adjusting electrical conductance to represent changing weights.

When multiple perceptrons are connected in layers, they form a multi-layer perceptron (MLP), capable of distinguishing non-linearly separable data. In an MLP, each neuron is fully connected to all neurons of both the preceding and succeeding layers. Additionally, each neuron includes a “bias” input, which provides a constant input to help shift the activation function. The concept of “bias” in this architectural context can be summarized as a constant term added to the weighted sum of inputs before applying the activation function, allowing the model to shift the decision boundary independently of the input values. This description reflects the standard, widely accepted structure of modern multi-layer perceptrons (MLPs) used in today's neural networks.

It is important to distinguish this architectural bias from the statistical concept of the bias-variance tradeoff~\cite{belkin2019biasvariance}. The latter refers to a model's tendency to either oversimplify patterns (high bias, underfitting) or overfit to noise (high variance, overfitting). While both concepts relate to model performance, they address fundamentally different aspects of learning.In modern deep learning architectures, such as those powering large language models (LLMs), bias terms remain a fundamental component. They ensure that networks can learn complex representations by shifting activation thresholds, improving convergence during training, and enhancing the model’s ability to generalize across diverse inputs.

Evidently, “bias” is a technical term in AI development, particularly in machine learning, where it refers to a constant input term that shifts the decision boundary of a model to aid learning and improve convergence. This architectural definition is distinct from the statistical concept in the bias-variance tradeoff and should not be confused with societal concerns related to “salient social categories”~\cite{bender2021stochastic} addressed in bias-mitigation strategies.

\section{What 'Bias' Means in Everyday Language: Technical Term or Social Concept?}
In current academic discourse, the term “bias”, when used in discussions surrounding AI, is typically more associated with concepts like “prejudiced” or “bigoted”; the term conjures associations with discriminatory ideology such as racism, sexism, or homophobia. Oftentimes, in discourse unrelated to AI, this is what people are trying to describe when using the word bias. However, it is important to remember that the term can also be used to mean something more like “preference”. Especially in everyday social contexts, 'bias' can imply prejudice, but it is often used more neutrally to express a personal inclination, preference, or even strong liking~\cite{newton2024purpleocean}. If a loved one tells someone their new haircut looks good, they might respond with “well, you’re a bit biased”.
Since “bias” in AI is clearly taken to mean something negative, its colloquial use is not a contender for defining what it means to say that a model is biased.

As we have discussed, bias is used as a technical term in computer sciences, but also has a colloquial meaning. In addition, 'bias' has a technical meaning in social sciences, where it refers to: \begin{quote}
A tendency (either known or unknown) to prefer one thing over another that prevents objectivity, that influences understanding or outcomes in some way.\\
\hfill \cite{bell2015bias} \end{quote}
This definition aligns more closely with what people critique when discussing bias-mitigation strategies and fairness in AI systems. 

Now that we understand what people mean when they talk about bias in AI, let’s explore it in more detail.

\section{Unbiased is not the same as fair}

Defining bias and fairness as the two extremes a model might fall between is, ultimately, a false dichotomy. The juxtaposition suggests that if a model is fair, that means it’s unbiased. Let us have a look at this proposition for a moment and attempt to find an ex-negativo definition of what is meant by "bias" by firstly looking at what fairness means: \begin{quote}
Fairness is about equitable treatment that accounts for different circumstances and needs. While equality means providing everyone with the same resources or opportunities, fairness considers individual contexts and removes barriers that hinder participation or success.\\
\hfill\cite{oxfordreview2024fairness}
\end{quote}
But what, then, makes a model fair? In order to answer this question, we need to elucidate what is meant when asking for such a model. 
Consider, for example, sampling bias: 
\begin{quote}
Training data [that] is not representative of the population it serves [leads] to poor performance and biased predictions for certain groups.\\
\hfill\cite{ferrara2023fairness}
\end{quote}
A common example is a facial recognition algorithm trained predominantly on white individuals, resulting in significantly lower accuracy for people of color—a clear instance of sampling bias causing harmful outcomes~\cite{ferrara2023fairness}.  We can take this to imply that an unbiased model would perform the same on people of all ethnic backgrounds, i.e., treat everyone the same.

Let us examine a different example for sampling bias: Consider a model built to make predictions about children’s growth and development in order to design a new curriculum. Suppose you could survey every child in the world to acquire training data. Hypothetically, this must lead to an unbiased model. Using data from every child will make sure that no one group is over- or underrepresented, i.e., the data is representative. Any number of factors like biological, genetic and environmental conditions can affect developmental processes. So the hypothetical training data will use data that accurately reflects these differences (and their percentages across the population) to create the best curriculum.

Now consider these three children: 
\begin{enumerate}[noitemsep, topsep=0pt]
    \item An able-bodied intersex child born to farmers in rural Asia.
    \item A physically disabled boy born to university-educated parents in suburban Africa.
    \item A selectively mute girl born to a blue-collar single parent in urban Europe.
\end{enumerate}
These children will lead very different lives with very different influences on their development and unique challenges.

Although the model is unbiased, as it treats every child equally, it applies a one-size-fits-all standard without considering individual needs. This raises the question: Is treating everyone equally the same as treating everyone fairly?  Is it \textit{fair} to expect these children to learn the same things at the same time and with the same speed?

If a real-world application were possible, we assume that most people would likely consider  this model biased. Specifically \textit{against} children with e.g., disabilities or toward an imaginary norm of the “ideal child”. But why? There is no sampling bias, no algorithmic bias, no representation bias and no measurement bias as defined in~\cite{ferrara2023fairness}. 
We think it's because the problem here is \textit{not} bias. The \textit{real} problem here is discrimination. 

\section{discriminating between vs Discrimination}
\begin{figure}[h]
    \centering
    \includegraphics[width=0.9\columnwidth]{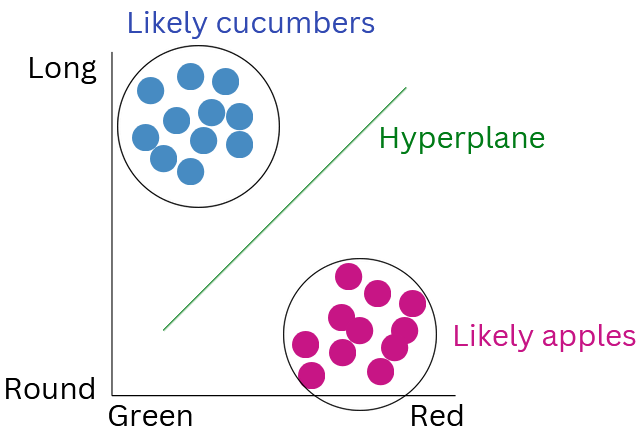}
    \caption{Linearly separated scatter plot showing differentiation between cucumbers and apples.}
    \label{fig:2}
\end{figure}

Let us, for a moment, return to the beginnings of AI, specifically to the MCPs and Perceptrons mentioned at the start of the paper. Their chief objective is to classify by distinguishing between categories based on measurable criteria. This technical form of discrimination—meaning differentiation—is value-neutral. No ethical concerns arise because the criteria are relevant to the task. Thus, this process does not lead to accusations of discrimination. 

But when AI is said to discriminate (which is, as shown above, what people actually try to express when saying it is biased), discussion around mitigation strategies based on obvious sentiment and arguments against discriminatory ideology that emphasize equal treatment for people irrespective of categories like ethnic background is brought up. Why is that?  
Simple neural networks like the ones mentioned previously aim to linearly separate data. If two things are linearly separable, that means it is possible to create a hyperplane in a two-dimensional vector space that has one category of thing on one side and another category of thing on the other. What parameters decide the distribution on either side? There are two dimensions (the axes) along which the data gets sorted. Say one axis depicts length to roundness and the other green to red. If that model is now fed pictures of cucumbers and apples, the result will be a clearly linearly separated scatter plot, a task any simple MLP can perform, see Figure~\ref{fig:2}. Importantly, for optic discrimination between apples and cucumbers, these are relevant criteria. The relevance of dimensions is crucial. Using length and color to distinguish apples from oranges would fail because those features do not meaningfully differentiate them. Failing to apply meaningful dimensions can be taken one step further: using evaluative labels rather than descriptive ones likely results in subjective, non-usable data. Relevance is what separates useful classification from arbitrary or harmful discrimination.
Overall, applying objective criteria that are relevant to a distinction is something generally associated with “discriminating between”. 

So what is “Capital D Discrimination”? Let us take another look at~\cite{ferrara2023fairness}: an example of algorithmic bias they give is this: 
\begin{quote}
An algorithm that prioritizes age or gender, leading to unfair hiring decisions. 
\end{quote}
Why is this unfair? Hiring decisions should generally be made with qualifications in mind. People consider it unfair when women have less chance of being hired or are hired for less money than men. But why? 

We propose that the answer to this question is that the difference between “discrimination between groups” and “Capital D Discrimination” is whether the distinction in question is made along relevant or irrelevant dimensions. 

Examples in academic literature for the impact of bias highlight various harmful outcomes, including:
\begin{quote}
harmful outcomes [relating to] racist, sexist, ableist, extremist or otherwise harmful ideologies [\ldots] reinforc[ing] hegemonic viewpoints [and] stereotypical associations or negative sentiment towards specific groups.\\ 
\hfill\cite{bender2021stochastic}.
\end{quote}

These examples highlight the sociological meaning of bias, which differs from technical bias. Technical bias refers to systematic errors in model predictions due to flawed data or algorithms, while sociological bias refers to preferential treatment that impacts outcomes. Discrimination, however, refers to harmful outcomes stemming from unjust social hierarchies and stereotypes. 

Seeing how the “bias” concern in AI resides on a sociological definition, we propose also looking at it's supposed opposite “fairness”, i.e. “un-biasedness”, from a sociological angle. Sources like~\cite{jardinez2024inclusive} demonstrate clearly that simply treating (and in this context teaching) everyone the same, which is what so-called bias mitigation strategies usually aim to do, does not ensure the best outcome for disadvantaged groups. On the contrary, only accommodation of differences like the ones outlined in \cite{jardinez2024inclusive} improves access and opportunities for marginalized groups. Therefore, treating everyone the same is not, in fact, fairness. True fairness, or equity, requires recognizing and accommodating differences to ensure that everyone has equal access to opportunities.
It follows from this that "bias" is not accurately categorised as oppositional to fairness. Based on a sociological definition of fairness striving for equity, it is more accurately juxtaposed with discrimination.

\section{Conclusion}
\begin{figure}[h]
    \centering
    \includegraphics[width=0.75\columnwidth]{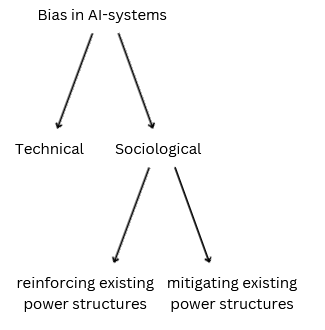}
    \caption{Flowchart illustrating discussed types of biases.}
    \label{fig:3}
\end{figure}

Our analysis demonstrates that the term "bias" is often used ambiguously, conflating technical, sociological and social meanings in a way that risks undermining efforts to address real harms.

As shown in Figure~\ref{fig:3}, we have identified two distinct categories of “bias”:
\begin{itemize}
    \item \textbf{Technical bias:} A neutral, architectural concept referring to input modulation in neural networks.
    \item \textbf{Social bias:} A value-laden concept concerning unequal treatment. It has two forms:
    \begin{itemize}
        \item \textbf{“Bad” bias (Capital D Discrimination):} Unfair treatment based on outcome-irrelevant characteristics (e.g., race or gender in hiring decisions).
        \item \textbf{“Good” bias (Equity):} Fair differentiation that acknowledges outcome-relevant differences (e.g., individual needs or abilities in a teaching context).
    \end{itemize}
\end{itemize}

Recognizing these distinctions is essential for developing effective bias mitigation strategies. Efforts that conflate technical and social bias or treat fairness merely as "equal treatment" risk perpetuating inequalities rather than correcting them.
In conclusion, fairness in AI cannot be reduced to the absence of bias. True fairness requires distinguishing between harmful discrimination and equitable differentiation—because sometimes, the fairest solution is to treat people differently.


\bibliography{main}




\end{document}